\documentclass{article}

\usepackage{arxiv}

\usepackage{amsmath,amsfonts}
\usepackage{algorithmic}
\usepackage{algorithm}
\usepackage{array}
\usepackage{textcomp}
\usepackage{stfloats}
\usepackage{hyperref}
\usepackage{verbatim}
\usepackage{graphicx, subfigure}
\usepackage{cite}
\usepackage[dvipsnames]{xcolor}
\usepackage{fancyhdr}

\usepackage{multirow}
\usepackage{float}
\usepackage{booktabs}
\usepackage{adjustbox}

\usepackage{tabularx}
\newcommand{\red}[1]{\textcolor{red}{#1}}

\usepackage{amssymb}
\usepackage{xcolor, soul}
\sethlcolor{green}
\usepackage[algo2e]{algorithm2e}

\begin{document}

%
\title{DeepTrack: Lightweight Deep Learning for Vehicle Trajectory Prediction in Highways}


\author{
    Vinit Katariya
    \thanks{Published in IEEE Transactions on Intelligent Transportation Systems. Available at: \href{https://doi.org/10.1109/TITS.2022.3172015}{https://doi.org/10.1109/TITS.2022.3172015}}\\
    The UNC at Charlotte\\
    Charlotte, NC\\
    \texttt{vkatariy@uncc.edu}\\
    \And
    Mohammadreza Baharani\\
    The UNC at Charlotte\\
    Charlotte, NC\\
    \texttt{mbaharan@uncc.edu}\\
    \And
    Nichole Morris\\
    University of Minnesota\\
    Minneapolis, MN\\
    \texttt{nlmorris@umn.edu}\\
    \And
    Omidreza Shoghli\\
    The UNC at Charlotte\\
    Charlotte, NC\\
    \texttt{oshoghli@uncc.edu}\\
    \And
    Hamed Tabkhi\\
    The UNC at Charlotte\\
    Charlotte, NC\\
    \texttt{htabkhiv@uncc.edu}\\
}


\markboth{Submitted to IEEE Transactions on Intelligent Transportation Systems}%
{Shell \MakeLowercase{\textit{et al.}}: Bare Demo of IEEEtran.cls for IEEE Transactions on Magnetics Journals}

\maketitle

%



\begin{abstract}
Vehicle trajectory prediction is essential for enabling safety-critical intelligent transportation systems (ITS) applications used in management and operations. While there have been some promising advances in the field, there is a need for modern deep learning algorithms that allow real-time trajectory prediction on embedded IoT devices. This article presents DeepTrack, a novel deep learning algorithm customized for real-time vehicle trajectory prediction and monitoring applications in arterial management, freeway management, traffic incident management, and work zone management for high-speed incoming traffic. In contrast to previous methods, the vehicle dynamics are encoded using Temporal Convolutional Networks (TCNs) to provide more robust time prediction with less computation. DeepTrack also uses depthwise convolution, which reduces the complexity of models compared to existing approaches in terms of model size and operations. Overall, our experimental results demonstrate that DeepTrack achieves comparable accuracy to state-of-the-art trajectory prediction models but with smaller model sizes and lower computational complexity, making it more suitable for real-world deployment.

\end{abstract}

\keywords{Vehicle trajectory prediction, Deep Learning, Temporal Convolutions, DeepTrack}

\section{Introduction}

With the advent of high-speed communication systems and unprecedented improvements in trajectory predicting, we are closer to implementing a fully connected (vehicle-to-vehicle (V2V) and vehicle-to-infrastructure (V2I)) and fully-aware transportation system than ever before. The increasing push towards autonomous driving and thrust for designing the best in class crash-avoidance systems at the edge has resulted in development of trajectory prediction algorithms with better than $90\%$ accuracy for up to 5 seconds in the future. The use of such high accuracy models in safety-critical systems for crash avoidance and accident prediction can result in precise time-to-collision (TTC) prediction \cite{lefevre2014survey}, which can prove instrumental in avoiding accident-related injuries and saving many lives.

Fast and effective prediction of future paths of surrounding vehicles and consequently making automated adaptive decisions will improve the safety and efficiency of autonomous vehicles and driving assistance systems, especially in complex and less predictable scenarios caused by major contributors to accidents. Agile and accurate trajectory predictions will also improve the decision-making of autonomous vehicles towards enhancing ride comfort, energy consumption, and traffic congestion \cite{phan2020covernet, lin2020attention}. In 2019, there were 36,096 fatalities on roadways in the United States \cite{nhtsa_fatality_NationalStats, nhtsa_fatality_VehiclesInvolved}. NHTSA (2019) also estimates that $11.9\%$ of them involved a vehicle maneuvering in a manner that may be unpredictable to the other drivers (i.e., turning left or right, stopping or slowing in traffic, merging/changing lanes, or passing another vehicle). Such crashes at highway speeds, given the short TTC and limited distance range, cannot be prevented with vision-based systems alone \cite{Horst1994TTC}.

Real-time trajectory prediction on-spot is a quite challenging task due to the uncertain and dynamic nature of roadways. We often observe many non-linearities on vehicle trajectories, stream from nearby vehicles or the number of available lanes (environmental factors), or individual decisions or preferences (intrinsic factors). Predicting multiple possible trajectories for an active subject in the scene \cite{deo2018convolutional, phan2020covernet} is a common practice. These trajectories are ranked based on the probability distribution of the prediction model, which makes them inherently less practical in real-time scenarios. Some recent approaches also consider the interactions of the nearby vehicles to successfully predict the path of a moving vehicle \cite{Hou2020InteractLSTM, deo2018convolutional}. However, they often have a relatively large model size and high computational complexity. A larger model size translates to higher storage requirements and higher system cost~\cite{Lei2019DataCollection}. Smaller model size translates to faster performance and low memory requirement. Data storage and compression mechanisms are also crucial to reducing the system size and cost in V2X infrastructure. 

This article proposes DeepTrack as a novel deep learning algorithm with comparable accuracy to best-in-class trajectory prediction algorithms but with a much smaller model size and lower computational complexity. DeepTrack encodes the vehicle dynamics with the aid of Temporal Convolutional Networks (TCN) instead of well-established mainstream Long-Short-Term Memory (LSTM) units~\cite{DBLP:journals/neco/HochreiterS97} . TCN, with its depthwise convolution as its backbone, can shrink the complexity of models and boost gradient flow for a more generalized trained model compared to LSTM-based solutions. We also augment DeepTrack with time attention modules to enhance the robustness against the noise and provide higher accuracy with minimum computational overhead. For the experimental results and comparison, this article uses datasets provided by the Federal Highway Administration (FHWA) under Next Generation Simulation (NGSIM) program.

Compared to CS-LSTM \cite{phan2020covernet}, DeepTrack reduces Average Displacement Error (ADE) by 12.23\%, Final Displacement Error (FDE) by 2.69\%, and also reduces the number of operations and model size by about 21.67\% and 43.13\%, respectively. CF-LSTM \cite{xie2021congestion} is outperformed by DeepTrack by 2.43\% in terms of ADE while operations and model sizes are reduced by 22.37\% and 43.75\%, respectively. The DeepTrack model reduces the number of operations and model size by 22.84\% and 12.61\%, respectively, over STA-LSTM \cite{lin2020attention}, while ADE is lower by 5.97\% (around maximum of 15 cm) and FDE is lower by 2.77\%.

Overall, the key contributions of this article are:

\begin{itemize}
    \item A lightweight trajectory forecasting model to precisely predict the location of vehicle of interest up to 5 sec. in the future.
    
    \item A novel encoder design based to reduce the complexity and size of the overall network by at least 22\% and 12\% compared to state-of-the-art trajectory prediction models with comparable accuracy.

    \item An extensive analysis on design of proposed network highlighting the effect of various components on the performance.
\end{itemize}



\section{Related Work}
\label{sec:relateWorks}
Vehicle trajectory prediction networks are into three types, Physical-based, Maneuver-based, and Interaction-aware models \cite{lefevre2014survey}. The physical-based models \cite{amm2009Physics, polychronopoulos2007sensor, schubert2008comparison} are designed using the laws of physics, maneuver-based models \cite{liu2014estimation, althoff2009model} consider the driver intentions, and interaction-aware models \cite{mo2020interaction, deo2018convolutional, ju2020interaction} take surrounding vehicles and their interactions into account for motion and path prediction. Compared to traditional physical-based models and maneuver-based models, which have limited visibility and thus accuracy, interaction-aware models can achieve much better accuracies. In past, interaction conscious networks used to be Dynamic Bayesian \cite{kafer2010recognition,lawitzky2013interactive} or prototype trajectory models \cite{brand1997coupled, 9492909}. With invent of the deep learning paradigm, many recent works use Long Short-Term Memory (LSTM) neural networks \cite{wang2020knowledgeLSTM, lin2020attention, xin2018intention} to capture the information of the neighboring vehicles.

The work by Deo $et~al.$ \cite{deo2018convolutional} combines LSTM encoding of the target vehicle with a Maneuver-based LSTM decoder to forecasts multiple trajectories based on maneuver classes. In \cite{mercat2020multi}, Mercat $et~al.$ also proposed LSTM based encoder-decoder architecture but avoid using predefined maneuver classes. It has self-attention layers in the middle that accept the encoded information of each vehicle for specific time instances. This helps in generating a fixed-sized input even when the number of vehicles in the scene might change. The performance of LSTM based model in \cite{mercat2020multi} is better than most of the best in class trajectory predicting algorithms. LSTM-based trajectory prediction have also been proposed in aircraft trajectory prediction~\cite{altche2017lstm} for higher positional reliability and safety. The work by Xie. $et~al.$ \cite{xie2021congestion} proposes a Graph Convolutional Neural Network (GNN) \cite{Xu2019GNN} based teacher-student model that predicts higher accuracy trajectories than the previous models. The teacher model accepts frame-wise graph input built to reflect the positions of all the agents in the input frame. The student model uses LSTM based encoder-decoder for trajectory prediction and matches the congestion pattern of the teacher model to improve the accuracy of prediction. 

DeepTrack uses TCN based novel encoder to grasp the positions of the vehicles in the scene as compared to LSTM encoders in modern models \cite{xie2021congestion, mercat2020multi}. The output of encoders will condense the details of vehicular interaction in the past using convolutional layers In the following subsection we discuss evolution and basic building blocks of a generic TCN architecture~\cite{bai2018empirical}.

Another important aspect of intelligent traffic systems~\cite{Zadobrischi2021IntelligentTraffic, Sabeti2021AIAR} using deep learning models on the edge devices and connected vehicles is balancing memory requirements with equipment cost. Lin et. al. \cite{Lei2019DataCollection} discuss the importance of saving data storage using compression techniques \cite{Muckell2014TrajComp}. It is also shown that this can help in improving the travel time estimation error up to 65\% along with reducing the on-board memory capacity. DeepTrack does not use message compression techniques. However, the model size and data storage capacity is directly related to overall equipment costs of the system~\cite{Chiu22cost}. A compact model such as Deep Track can help designers save storage and lower the system cost.

\section{Motivation: Applications of Real-time Trajectory Prediction in Highways}
\label{sec:motivation}
Predicting vehicle trajectories on the highway offers a variety of safety applications, particularly where points of conflict increase between road users. There are expected V2I applications to aid active traffic management systems to divert traffic away from predicted conflicts through lane-use control signals, dynamic hard shoulder running, and variable speed limits~\cite{abdel2008,Dutta201966TMS}. Leveraging predictions to calm traffic through speed reductions or divert upstream traffic away from high conflict zones would be expected to reduce the potential for other vehicles to be impacted by potential collisions or constrain traffic flow. Moreover, such predictions may also reduce the likelihood of "secondary crashes" occurring in the aftermath of primary crashes~\cite{YANG2014143}, which are estimated to account for up to 15\% of all freeway crashes~\cite{raub1997}. In congested freeway platooning conditions, free-flowing traffic is at risk of propagating increasing braking responses upstream due to downstream hard braking, close-call, or collision events. The increased braking responses, paired with slow response times, propagated upstream risk rear-end collisions. Providing real-time traffic control to upstream traffic based on downstream traffic predictions of safety-critical events may offer the potential to dampen the effects of traffic instability by reducing time lags in response~\cite{kesting2008howtoreact}.
    
Further, there are expected V2X applications which may reduce the risk of predicted vehicle trajectories into high-risk areas such as work zones or roadside emergency work areas by providing advanced alerting to on-duty workers to seek safety~\cite{Yo2013}. Emergency responders are at risk of being struck by vehicles while conducting traffic stops or attending to a crash. Often these crashes occur due to distracted or impaired drivers unintentionally leaving their lane of travel~\cite{tijerina2003committee}. Secondary crashes are also a risk for emergency responders when traffic is not properly controlled around and upstream from the scene~\cite{codd2014fatal}. Providing predictions of vehicle trajectories near the roadside, on-duty workers may also help to guide better traffic management through Portable Changeable Message Signs (PCMS) to alert drivers to risks in addition to emergency alerting to the workers themselves.

\section{DeepTrack}
\label{sec:deeptrack}

This section discusses the DeepTrack architecture in detail.

The inputs of DeepTrack are defined as:
\begin{gather}
    \boldsymbol{X}_{ego}=\left[
    \begin{array}{cccc}
    \label{eq:ego_input}
    l_{e}^{t_{-h}} & l_{e}^{t_{1-h}} & \cdots & l_{e}^{t_{0}}
    \end{array}
    \right],\\
    \boldsymbol{X}_{nbr}=\left[
        \begin{array}{cccc}
        \label{eq:nbr_input}
        l_{0}^{t_{-h}} & l_{0}^{t_{1-h}} & \cdots & l_{0}^{t_{0}} \\
        \vdots & \vdots& \ddots & \vdots \\
        l_{j}^{t_{-h}} & l_{j}^{t_{1-h}} & \cdots & l_{j}^{t_{0}}\\
        \vdots & \vdots& \ddots & \vdots \\
        l_{N-1}^{t_{-h}} & l_{N-1}^{t_{1-h}} & \cdots & l_{N-1}^{t_{0}}
    \end{array}
    \right], 
%
\end{gather}
where $l_{i}^{t}=\left<x_{i}^{t}, y_{i}^{t}\right>$ is the position of vehicle $i$ at time $t$, $h$ is number of seconds of the past trajectory used for prediction, $\boldsymbol{X}_{e}$ is the car of interest, $\boldsymbol{X}_{nbr}$ is neighbour cars, and $N$ is the number of vehicles. The shape of $\boldsymbol{X}_{nbr}$ is shown in Fig.~\ref{fig:atcn}. In a similar way, the output can be defined as follows:

\begin{equation}
\label{eq:output}
    \boldsymbol{\hat{Y}}=\left[
        l^{t_{1}}, l^{t_{2}}, \cdots, l^{t_{f}}
    \right].
\end{equation}


\subsection{Model Architecture}
We present the proposed DeepTrack architecture in Fig.~\ref{fig:bd}. It consists of an encoder, a Vehicular Interactive Aware Convolution (VIAC), and an LSTM trajectory encoder. In the following sections, we explain the design and working of each component.

\subsubsection{DeepTrack Encoder}

\begin{figure}[!b]
	\centering
	\vspace{-20pt}
	\includegraphics[width=0.9\linewidth,trim= 10 10 2 10,clip]{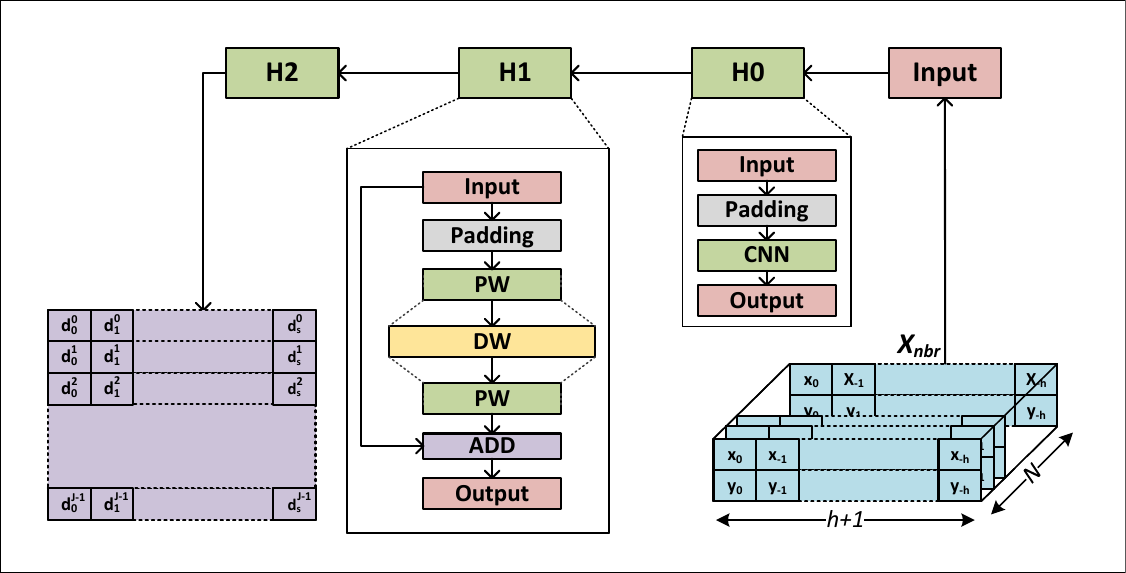}
	\vspace{-10pt}
	\caption{Structure of TCN based DeepTrack encoder.}
	\label{fig:atcn}
\end{figure}

Inspired by the generic TCN architecture, we propose an encoder to reduce model complexity and memory footprint. Fig.~\ref{fig:atcn} shows the structure of a single DeepTrack encoder block. The DeepTrack encoders embed vehicles' path histories, both neighbors and target vehicles, into higher dimensions to capture their past trajectory. As opposed to previous works \cite{xie2021congestion, deo2018convolutional}, DeepTrack does not require dense layers to embed input features as needed for the LSTM encoder. Encoder convolutional operators capture and map the sequence of $l^{t}$ by applying $K=\left[ 2, k\right]$ as a kernel, where $k\in \mathbb{N} | k \leq h$. As a result, DeepTrack has less model complexity, and better gradients flow from output to input during optimization.

\subsection{Preliminaries and Problem Formulation}

Traditional Convolutional Neural Networks (CNN) are used in computer vision applications due to their success in capturing spatial features within a two-dimensional frame. Recently, research has shown that specialized CNNs can recognize patterns in data history to predict future observations. This gives researchers interested in time-series forecasting options to choose other options over RNNs~\cite{DBLP:journals/corr/ChungGCB14,DBLP:journals/neco/HochreiterS97}, an established DNN for time-series predictions. In one such case, TCN achieved the state of the art accuracy in sequence tasks, e.g., polyphonic music modeling, word and character-level language modeling, and audio synthesis \cite{oord2016wavenet,  DBLP:convolutionalencoder, DBLP:convseq2seq}.

TCNs are designed around two basic principles: 1) the convolutional operations are causal, i.e., predictions are made based only on current and past information; 2) the network receives an input sequence of arbitrary length and maps it to an output sequence of the same length. The use of causal convolutions in WaveNets~\cite{bai2018empirical} showed that it allowed for faster training as compared to LSTM based networks as they do not rely on recurrent connections. However, as the causal convolution needs large number of layers to increase the receptive field, WaveNet uses dilated convolutions to address this problem. In Dilated convolutions, the kernel is stretched to cover a larger part of the input. This is achieved by inserting holes (zeros) between the kernel elements. The level of enlargement is determined by dilation rate, which defines the number of spaces inserted between the kernel elements. Generally, d-1 spaces are inserted for dilation rate of d.



Simple causal convolutions have a dilation rate of 1, but other researchers incorporate dilated convolutions to scale the receptive field exponentially. The dilated convolution of $F$ on element $s$ of a sequence $X$ is given as:
\begin{gather}
\label{eq:dilated}
F(s) = (x*_{d}f)(s) = \sum_{i=0}^{k-1}f(i)\cdot x_{s-d \cdot i},
\end{gather}
where $X \in \mathbb{R}^{n}$ is a 1-D input sequence, $*_{d}$ is dilated convolution operator, $f : \{0, ..., k-1\} \in \mathbb{R}$ is a kernel of size $k$ and $d$ is the dilation rate \cite{bai2018empirical}. Also receptive filed of a dilated convolution can be calculated by:
\begin{equation}
\label{eq:receptiveField}
rf = 1 + \sum_{j=1}^{L}[k(l)-1]\times d(l),
\end{equation}
where $j \in \{1, 2, 3, ..., L\}$ is the layers, $k$ is the kernel size, and $d(j)$ is the dilation rate at layer $j$. This means that as the depth of the network increase, so does the receptive field. To address the issue of vanishing gradients resulting in exponentially expanding receptive fields with increasing network depth, TCN replaces standard convolutional layer in the residual block ~\cite{DBLP:conf/cvpr/HeZRS16}. This is a widely used approach for convolutional architectures  as it provides a path for information to pass  through the layers. A residual block can be represented as:
\begin{equation}
\label{eq:residual}
y = \mathcal{F}_{i}(X)+X
\end{equation}
where, $y$ is the output of residual block, and $\mathcal{F}_i$ represents the operations such as, convolutions layers , non-linearity, and normalisation applied to input $X$ at each layer $i$. The residual block helps the network in learning the modifications applied to input at each layer \cite{bai2018empirical}.

\begin{figure*}[t]
\centering
\includegraphics[width=.85\linewidth, trim= 25 20 25 20,clip, keepaspectratio]{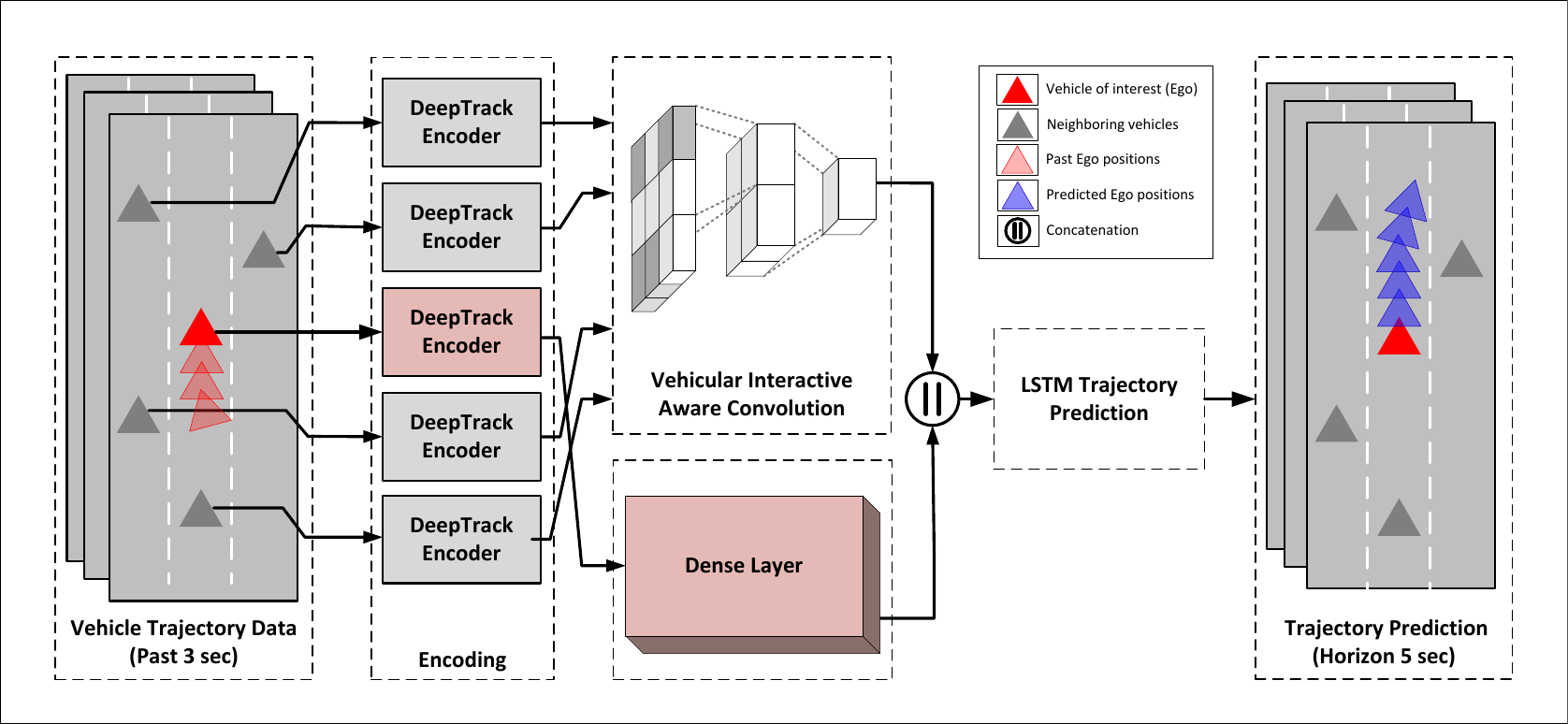}
\caption{Overview of the trajectory prediction model. The location of the neighbors (gray triangles) and car of interest (solid red triangle) is shown at $t_{0}$ in Vehicle trajectory data (extreme left) block and Trajectory prediction (extreme right) block. Triangles denoting semi-transparent red in Vehicle trajectory data block, and semi-transparent blue in Trajectory prediction block represent observed history paths, and model output respectively. The observed history paths of neighbours (for past 3 seconds) are used by the model but not shown in the figure to avoid confusion.}
\label{fig:bd}
\vspace{-10pt}
\end{figure*}

DeepTrack uses two different encoders. There is one shared by all neighbors, shown by the shaded box in Fig.~\ref{fig:bd}, which maps their dynamics to higher dimensions so that the VIAC can comprehend their interdependencies.  The other encoder maps only the ego dynamics, as illustrated by the red box in Fig.~\ref{fig:bd}.

Since we have a padding unit in each encoder to make sure the input and output of standard and depthwise convolution will be the same (second principle of TCN architectures), the $2p$ zeros are added symmetrically, where $p$ is given by:
\begin{gather}
    p = \lceil \frac{(o-1)\times s+(k-1)\times(d-1) -i+k }{2} \rceil,
    \label{eq:padding}
\end{gather}
where $o$ is the output size, $i$ is the input size, $s$ is the stride, $k$ is the kernel, and $d$ is the dilation.  According to Eq. \ref{eq:padding}, if we increase the kernel size or dilation, more zeros should be padded to the input. The addition of excessive zeros to the input has two main disadvantages: \textcircled{\raisebox{-0.9pt}{1}} it degrades the model's performance due to redundant zeros, and \textcircled{\raisebox{-0.9pt}{2}} it increases the model computational complexity. As a result, we set the dilation and kernel size of the DeepTrack encoders for different models as shown in Table \ref{tab:atcn_encoder}, both DeepTrack encoders have three hidden layers; however, the output dimensions differ.



Each DeepTrack encoder has a three-layered structure with a padding block to ensure that the input and output sizes are the same, followed by a convolutional block. Layer H0 uses standard convolution ($d = 1$). As a widely accepted practice in the deep learning community, batch normalization \cite{DBLP:conf/icml/IoffeS15} layer is added after each convolution to speed up and stabilize the model training followed by a ReLU activation. We have intentionally not shown Batch Normalization (BN) and ReLU activation in Fig. \ref{fig:atcn} for simplification of the diagram.


For DeepTrack, Swish, as well as ReLU activation, were tested after BN. Swish is very similar to ReLU but does not abruptly change its direction. However, we use ReLU activation for DeepTrack to provide marginally better results. BN and ReLU are not shown in Fig.~\ref{fig:atcn} for simplification.

In the subsequent layers (H1-H2), a padding block is followed by a pointwise (PW), depthwise (DW), and PW convolutions instead of a CNN to reduce the model complexity. As Layers H1 and H2 are identical, details of H2 are not shown in Fig \ref{fig:atcn}. BN and ReLU activation is applied to the output of the last PW convolution before adding it to the input of the residual block~\cite{DBLP:conf/cvpr/HeZRS16} as expressed by Eq.~\ref{eq:residual}.

To the best of our knowledge, we are first to present a deep learning algorithm with a modified generic TCN architecture for trajectory prediction.

\subsubsection{Attention Mechanism}

Attention mechanisms have been used to better interpret the model and extract the significant information. It has also been used in multiple trajectory prediction applications as shown in~\cite{DBLP:journals/corr/abs-2007-03639, yan2020trajectory}.

\begin{figure}[!htbp]
	\centering
	\vspace{-10pt}
	\includegraphics[width=0.7\linewidth,trim= 10 10 10 10,clip]{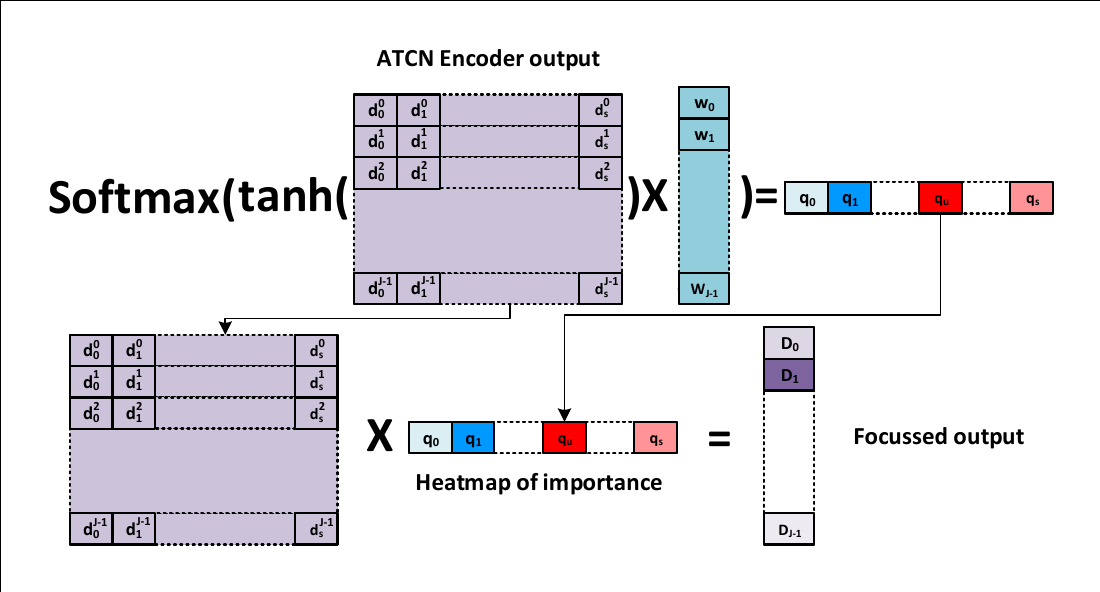}
	\caption{Additive Attention. The output of softmax is multiplied with DeepTrack encoder output to lead the next layers on essential features.}
	\label{fig:att_mech}
\end{figure}

In this architecture, additive attention mechanism \cite{additive_attention} for both DeepTrack encoders to guide decoders are used to predict the trajectory based on the importance of features, similar to the work by Lin $et~al.$ \cite{lin2020attention}. In order to get the importance of the encoded output, the first associated weight score vector, $\overrightarrow{\omega}$, should be calculated by:

\begin{gather}
\label{eq:att_eq1}
    \overrightarrow{\omega}=tanh(WD),\\
    \boldsymbol{D}=\left[
        \begin{array}{cccc}
        l_{0}^{t_{0}} & l_{0}^{t_{1}} & \cdots & l_{0}^{t_{s}} \\
        \vdots & \vdots& \ddots & \vdots \\
        l_{J-1}^{t_{0}} & l_{J-1}^{t_{1}} & \cdots & l_{J-1}^{t_{s}}
    \end{array}
    \right],
\end{gather}
where $D$ is the DeepTrack encoder output, $s$ is the encoder output length size set to $h$, $J$ is the last output channel size, and $W$ is trainable parameter. The final attention score is then given by:
\begin{gather}
\label{eq:att_eq2}
    f_{att}=\sigma(\overrightarrow{\omega}),\\
    D_{final}=D^{T}f_{att},
\end{gather}
where $\sigma$ is $Softmax$ function. Fig.~\ref{fig:att_mech} illustrates the attention mechanism. The output of $softmax$, $f_{att}$ is the importance heatmap, and it will be again multiplied with the encoded data. Based on the focused data, the VIAC and LSTM trajectory decoder can figure out the vehicle interactions and generate the final prediction effectively.

\subsubsection{VIAC}
\label{sec:viac}
Analyzing the interaction between the ego and its neighbors is necessary to predict the future trajectory for the vehicle of interest. Despite capturing individual behavior, the DeepTrack encoder is unable to comprehend the entire scene. Social pooling proposes a solution by pooling encoded data around a specific target \cite{DBLP:conf/cvpr/AlahiGRRLS16}. The task is accomplished by defining a spatially correlated grid $f \times g \times N$ regarding the car of interest. Similar to Social Pooling \cite{deo2018convolutional}, we set $f$ and $g$ to 13 and 3, respectively.
The structure of the $3\times3$ grid that masks the encoder output is shown in the VIAC block of Fig. \ref{fig:bd}.

DeepTrack comprehends the interdependencies of the vehicle by applying convolutions to embed information. The neighborhood dynamics are encoded and mapped to the lower dimension using the two-layer convolution and a pooling unit. The convolutional layers help extract the local features from the spatial grid around the ego vehicle. The use of convolutional-social pooling in the VIAC shows lower performance degradation as compared to fully connected social pooling, as shown by Deo $et.~al$ in ~\cite{deo2018convolutional}.

The additional dense layer dedicated to the dynamic encoding of the ego vehicle is fully connected. The dense layer also remaps the DeepTrack encoder output of ego to have the same feature size so that it can be concatenated with the result of VIAC as shown in Fig.~\ref{fig:bd}. The concatenated output is then passed through the LSTM Trajectory Prediction block.

\subsubsection{LSTM Trajectory Prediction}
Only at the final stage, DeepTrack uses an LSTM-based encoder to predict the future trajectory, $\hat{Y}$. We have not used TCN based encoder as the final stage because TCN can map the temporal information only to a higher output channel. Similar to what is accomplished by the DeepTrack encoder at the first stage. LSTM is only used to map and decode the VIAC and Dense layer's concatenated output to the final output prediction.

\subsection{Algorithm}
\label{sec:algorithm}
Algorithm \ref{algo:deeptrack} represents a step-by-step working of the DeepTrack prediction network. Input $X_{nbr}$ has dimension $(N-1) \times (h+1)$, and $X_{ego}$ has dimension $1\times (h+1)$. Expected output, $\hat{Y}$, is given by eq.~\ref{eq:output}.

\subsubsection{Encoder Functions}
There are two encoder functions in Deeptrack, $Encoder_{nbr}$ and $Encoder_{ego}.$ Both Encoder functions have three hidden layers represented by blocks H0, H1, and H2 in Fig.~\ref{fig:atcn}. H0 uses standard convolution, but H1 and H2 use point-wise, depth-wise, point-wise convolutions. All the convolution operations are followed by appropriate padding, normalization, and activation. Number feature sizes for each block are as shown in Table \ref{tab:atcn_encoder}.

First, $Encoder_{nbr}$ and $Encoder_{ego}$ functions is applied to input vectors $X_{nbr}$ and $X_{ego}$ respectively. Next, to incorporate the attention mechanism, Softmax function is applied to tanh activation of the dot product of $Encoder$ outputs $D_{nbr}$, and $D_{ego}$ with $W_{nbr}$, and $W_{ego}$ respectively. $W$ represents trainable attention weight vectors, and $\sigma$ represents $Softmax$ function. The attention mechanism is a part of encoder blocks in Fig. \ref{fig:bd}. The output of attention block is then passed to VIAC.

\subsubsection{VIAC, Dense Layer and $LSTM_{decoder}$ Functions}
VIAC and Dense Layer (DS) functions are applied to the dot product of the transposed output of $Encoder$ functions and attention mechanism for neighbor and ego vehicles. VIAC first combines the tensors for every grid with a car to form a single tensor. Next, a couple of convolutional layers are applied to this tensor, followed by a pooling layer. Simultaneously, the decoded state tensor of the ego vehicle is passed through a fully connected layer represented by $DS$ \cite{deo2018convolutional}. $DS$ translates the inputs into a feature size that is concatenated with VIAC output to produce a comprehensive encoded trajectory, $Y_{final}$. Concatenation is represented by `$||$' in the algorithm.

Finally, $LSTM_{decoder}$ function is applied to $Y_{final}$. The decoder has two softmax layers with outputs concatenated to predict the final trajectory of the ego vehicle, $\hat{Y}$. 

\begin{algorithm}[H]

 \SetKwData{Left}{left}\SetKwData{This}{this}\SetKwData{Up}{up}
\SetKwFunction{Union}{Union}\SetKwFunction{FindCompress}{FindCompress}

\SetKwInOut{Output}{Output}

\SetKwInOut{Input}{Input Neighbour}
  \Input{$X_{nbr}$ = all rows except row $l$ from metrics in eq.~\ref{eq:nbr_input}}

\SetKwInOut{Input}{Input Ego}
 \Input{$X_{ego}$ = row $l$ from metrics in eq.~\ref{eq:ego_input}.}
 
 \Output{$\hat{Y}$ = output as shown in eq.~\ref{eq:output} }
 Initialize encoder layers, kernels, dilation 
 
 $D_{nbr}$ = $Encoder_{nbr}$($X_{nbr}$)
 
 $f_{att\_nbr} = \sigma(tanh(W_{nbr}D_{nbr}))$
 
 $y_{nbr} = VIAC(D_{nbr}^T f_{att\_nbr})$
 
 $D_{ego}$ = $Encoder_{ego}$($X_{ego}$)
 
 $f_{att\_ego} = \sigma(tanh(W_{ego}D_{ego}))$
 
 $y_{ego} = DS(D_{ego}^T f_{att\_ego})$

$y_{final} = y_{ego}~ || ~y_{nbr}$
 
 $\hat{Y}$ = $LSTM_{Decoder}$($y_{final}$)

 \caption{DeepTrack prediction model pseudo code}
 \label{algo:deeptrack}
\end{algorithm}

\begin{table*}[b]
  \centering
  \caption{DeepTrack Variants}
  \begin{tabular}{cccccccc}
    \toprule
    \textbf{Models} & ${DT_0^*}$ & ${DT_1}$ & ${DT_2}$ & ${DT_3}$ & ${DT_4}$ & ${DT_5}$ & ${DT_6}$ \\
    \midrule
    \textbf{Features} & [16, 32, 64] & [32, 16, 64] & [32, 16, 64] & [32, 16, 64] & [32, 16, 64] & [16, 32, 64] & [16, 32, 64] \\
    \textbf{Convolution} & Separated & Separated & Separated & Normal & Normal & Normal & Normal \\
    \textbf{Activation} & Swish & ReLU6 & Swish & Swish & Swish & Swish & ReLU6 \\
    \textbf{Optimizer} & ADAM  & ADAM  & ADAM  & SGD & ADAM  & ADAM  & ADAM \\
    \bottomrule
    \end{tabular}%
  \label{tab:dt_variants}%
\end{table*}%

\vspace{-10pt}
\section{Evaluation}
\label{sec:exp}
The performance of DeepTrack is evaluated using NGSIM's widely used I-80 \cite{NGSIM_i80} and US-101 \cite{NGSIM_US101} vehicle trajectory datasets. A rate of 10 Hz is used for sampling the vehicle's trajectory for 45 minutes. Each dataset includes three segments of 15 minutes long of mild, moderate, and congested traffic. Similarly to \cite{xie2021congestion, mercat2020multi, lin2020attention}, we divided the dataset into three parts: training, validation, and testing. The dataset provides around 8 million data entries divided into 70\% training data, 10\% validation data, and 20\% test data. As discussed in CS-LSTM \cite{deo2018convolutional}, DeepTrack also uses a stationary frame of reference.

The direction of motion of the vehicles is defined by the head of the triangle as shown in Fig. \ref{fig:bd}. Lanes immediately next to the ego vehicle are considered for tracking the neighbors' position. This helps capture the effect of movement of immediate neighbors on the ego vehicle as they have maximum influence on its future trajectory. The area around the ego vehicle is converted into a 13$\times$3 size grid, with each grid cell 15 feet long and width equal to lane width. The ego vehicle is assigned the center cell, whereas each neighboring vehicle in the 13$\times$3 grid is assigned a cell based on the position of its front bumper around the ego.

Based on the work of \cite{deo2018convolutional}, each trajectory is also segmented into 8 seconds, where the first three seconds are used as a path that was observed, and the model will predict the following five seconds. Previous works downsampled each second by two \cite{ xie2021congestion, mercat2020multi}  to reduce the complexity of the LSTM encoder. While we are not limited to this fact, we also downsampled the inputs for a fair comparison. In the following subsections, we discuss the DeepTrack implementation environment, the effect of various components on the design, quantitative results by comparing with contemporary models, and qualitative prediction analysis for various scenarios. All models proposed in this paper are implemented using the PyTorch package, an open-source machine learning library. The training was performed on Nvidia Tesla V100 GPU for 30 epochs using the ADAM optimizer with a default learning rate of 0.001.

\subsection{Evaluation Metrics}
A comparison of DeepTrack against off-the-shelf algorithms on the NGSIM dataset was conducted to provide a comprehensive comparison. Root Mean Square Error (RMSE), Average Displacement Error (ADE), and, Final Displacement Error (FDE) are used as a measure of prediction accuracy and performance of the system. As DeepTrack is designed with edge-based real-time applications in mind, the number of MACs and parameters of the models are also compared to state-of-the-art models to present a perspective on model complexities. The RMSE at time $t$ is given by:
\begin{equation}
    RMSE^{t}=\sqrt{\frac{1}{N}\sum_{i=1}^{N}{(Y_{i}^{t}-\hat{Y}_{i}^{t})}^{2}},
\end{equation}
where $Y$ is the ground truth, $\hat{Y}$ is predicted output, and $N$ is the number of samples. Average Displacement Error (ADE) and Final Displacement Error (FDE) are also calculated to compare the average RMSE over 5 seconds and error in the final predicted position.

ADE refers to the mean square error (MSE) overall estimated points of every trajectory and the actual points, and FDE is the root mean square error distance between the final predicted trajectory points and ground truth. These evaluation parameters are used in the following sections to compare and analyze the performance of DeepTrack. 


\vspace{-10pt}
\subsection{DeepTrack Models and Comparisons}
This section helps in understanding the impact of various components on the performance of DeepTrack. Different variants of DeepTrack were designed and evaluated to analyze the influence of each alteration on the network's overall performance.

\subsubsection{Variant Models Design}
The variant models were designed by changing or removing one of the following components from the network: attention-mechanism, optimizer, activation function, convolution type, and neighbors output feature sizes. The attention mechanism helps in grasping the effect of the neighboring vehicles on the ego trajectory. A DeepTrack model without an attention mechanism was designed, and the results are discussed here. In this study, ADAM and Stochastic gradient descent (SGD) optimizers were utilized to adjust model parameters to reduce the training loss. Swish and Rectified Linear Unit (ReLU) were used as an activation function for all the layers in DeepTrack analysis. Separated and normal convolution-based networks were designed and tested to reduce the complexity and compare performances of the network. Table~\ref{tab:dt_variants} shows seven different variants of DeepTrack each varying from other in at-least one of the above mentioned aspects. The model with superscript $^*$, e.g., $DT_0^*$,  denotes an absence of the attention-mechanism.

\begin{table}[htbp]
  \centering
  \caption{Configuration of DeepTrack encoder for ego and neighbors}
  \begin{adjustbox}{width=1.\linewidth,center}

    \begin{tabular}{clccc}
    \toprule
    \multicolumn{2}{c}{\multirow{2}[4]{*}{\textbf{Variant Encoder}}} & \multicolumn{3}{c}{\textbf{Configurations}} \\
\cmidrule{3-5}    \multicolumn{2}{c}{} & \textbf{Output feature size} & \textbf{Dilation rate} & \textbf{Kernel size} \\
    \midrule
    \multirow{2}[1]{*}{\textbf{$DT_0^*$}} & Neigbours & [16, 32, 64] & [1, 1, 1] & [2, 2, 2] \\
          & Ego   & [8, 16, 32] & [1, 1, 1] & [2, 2, 2] \\
          \midrule
    \multirow{2}[1]{*}{\textbf{$DT_i$ }} & Neigbours & [32, 16, 64] & [1, 2, 4] & [8, 4, 2] \\
          & Ego   & [32, 16, 64] & [1, 2, 4] & [8, 4, 2] \\
    \bottomrule
    \end{tabular}%
    \end{adjustbox}
  
  \label{tab:atcn_encoder}%
\end{table}%

\begin{table*}[b]
  \centering
  \caption{Performance comparison of DeepTrack variants based on Root mean square error, Displacement errors and Computation.}
    \begin{tabular}{p{4.055em}ccccccccc}
    \toprule
    \multirow{2}[4]{*}{\textbf{Model}} & \multicolumn{5}{c}{\textbf{RMSE (m)}} & \multicolumn{1}{c}{\multirow{2}[4]{*}{\textbf{FDE (m)}}} & \multicolumn{1}{c}{\multirow{2}[4]{*}{\textbf{ADE (m)}}} & \multicolumn{2}{c}{\textbf{Complexity}} \\
\cmidrule{2-6}\cmidrule{9-10}    \multicolumn{1}{c}{} & \multicolumn{1}{c}{\textbf{1s}} &
\multicolumn{1}{c}{\textbf{2s}} &
\multicolumn{1}{c}{\textbf{3s}} & 
\multicolumn{1}{c}{\textbf{4s}} & 
\multicolumn{1}{c}{\textbf{5s}} &       &       & \multicolumn{1}{c}{\textbf{MACs}} & \multicolumn{1}{c}{\textbf{Parameters}} \\
    \midrule
    $DT_0^*$ & 0.45  & 1.13  & 1.90   & 2.84  & 4.03  & 3.34  & 2.07  & 2,917,994 & 171,703 \\
    $DT_1$ & 0.46  & \textbf{1.07}  & 1.84  & 2.78  & 3.93  & 3.23  & 2.02  & \textbf{2,804,419} & \textbf{109,099} \\
    $DT_2$ & 0.47  & 1.08  & \textbf{1.83}  & \textbf{2.75}  & \textbf{3.89}  & 3.25  & \textbf{2.01}  & \textbf{2,804,419} & \textbf{109,099} \\
    $DT_3$ & \textbf{0.44}  & 1.14  & 1.92  & 2.86  & 4.01     & 3.27  & 2.07  & 3,221,478 & 125,923 \\
    $DT_4$ & 0.46  & 1.12  & 1.89  & 2.82  & 3.96  & 3.24  & 2.05  & 3,030,408 & 118,755 \\
    $DT_5$ & 0.46  & 1.08  & 1.84  & {2.76}   & {3.90}  & \textbf{3.21}     & \textbf{2.01}  & 4,148,894 & 125,923 \\
    $DT_6$ & 0.45  & 1.09  & 1.85  & 2.77  & 3.9   & \textbf{3.21}  & \textbf{2.01}  & 3,221,478 & 125,923 \\
    \bottomrule
    \end{tabular}%
  \label{tab:variant_performance}%
\end{table*}%

As a part of the ablation study, several different designs of TCN-based trajectory prediction networks were studied. However, only seven models based on their effect on the overall system's performance are presented in this study. First three models $DT_0^*$, $DT_1$, and $DT_2$ use separated convolution with a combination of different activation functions, Attention-mechanism layer, and neighbour output features. Next four models $DT_3$ $DT_4$, $DT_5$, and $DT_6$ used normal convolution with a combination of other parameters. All the models were trained with a data split similar to \cite{deo2018convolutional}. The hidden layers, dilation rate and kernel sizes of various model encoders are shown in Table~\ref{tab:atcn_encoder}, $DT_i$ represents models $DT_1$ to $DT_6$ as the encoder parameters are fixed models with attention mechanism to limit excessive padding.

\subsubsection{Performance Comparison and Model Complexities}
Table~\ref{tab:variant_performance} shows the performance of DeepTrack variants in three areas, mean error at the end of each second in meters (RMSE) , final and average displacement errors calculated in meters (FDE, ADE), and model complexity (number of MACs and Parameters).

$DT_2$ and $DT_5$ have the best performance in terms of RMSE. $DT_2$, $DT_5$ and $DT_6$ have best ADE values, while $DT_5$ and $DT_6$ have the best performance in terms of FDE. There is only a 1.25\% (0.04 m) difference in the FDE of $DT_2$ and the best performing models. Models $DT_1$ and $DT_2$ prove to be best in complexity-based performance, which is one of the most important aspects of this study. Hence, it can be concluded that $DT_2$ with attention mechanism, ADAM optimizer, ReLU activation, separated convolutions, and output features of [32, 16,64] is has the best overall performance among all the DeepTrack variants considered in this study.

The model with SGD optimizer, Swish activation, and standard convolution, $DT_3$, and one without attention mechanism, $DT_0^*$,  have the worst RMSE except for the first second. $DT_0^*$ has the worst FDE and ADE performance, making a solid case for the use of ADAM optimizer. $DT_5$ has the worst performance in terms of complexity with 32.4\% higher MAC count than $DT_2$ and a joint highest in number of Parameters with $DT_6$. Other models have similar performances in terms of RMSE, FDE, and ADE, but the difference can be observed when the complexity of the algorithms is analyzed. As expected, the models using standard convolutions show higher complexity when the number of multiply-and-accumulates (MACs) and model parameters are compared. The number of MACs and parameters are lowest for $DT_1$ and $DT_2$ as they use separated convolutions resulting in lower complexity than all the models using standard convolution.

It can be concluded that using separated convolutions helps in reducing the complexity of a model. Comparison of models $DT_1$ and $DT_4$ shows that the introduction of separated convolution helps reduce the number of MACs by 7.5\% and the number of parameters by 8.1\%. The attention mechanism also helps improve overall performance, as shown in the comparison of $DT_0^*$ and $DT_1$. The use of ADAM optimizer is also justified by analyzing the mediocre performance of $DT_3$ with SGD optimizer. The analysis of effect of various factors on performance of DeepTrack continues as we present the effect of different amount of training data on the model performance in the next part.

\subsubsection{Generalization Study}
Table~\ref{tab:DataSplit} summarises the error-based performance of $DT_2$ as it is the best performing DeepTrack variant. Column 1 in table~\ref{tab:DataSplit} shows the data split ratios as Tr:Val:Ts representing train : validation : test set ratios respectively. The data-split of 70:10:20 is same as used in~\cite{deo2018convolutional} for fair comparison with other models discussed in next section.

The results of data split of 80:10:10 as compared to 70:10:20 shows a possibility of over-fitting as the error is higher for larger training dataset. Other datasets with 60\% and 50\% training data-split also show an increase in root mean and displacement errors for $DT_2$.

\begin{table}[htpb]
  \centering
  \caption{Performance comparison of best DeepTrack variant $DT_2$ based on different data splits of NGSIM dataset. RMSE, FDE, ADE are in meters.}
    \begin{tabular}{cccccccc}
    \toprule
    \multicolumn{1}{c}{\textbf{Dataset split}} & \multicolumn{5}{c}{\textbf{RMSE horizon}} & \multicolumn{1}{c}{\multirow{2}[4]{*}{\textbf{FDE}}} & \multicolumn{1}{c}{\multirow{2}[4]{*}{\textbf{ADE}}} \\
\cmidrule{1-6}    \multicolumn{1}{c}{\textbf{Tr:Val:Ts}} & \multicolumn{1}{c}{\textbf{1s}} 
& \multicolumn{1}{c}{\textbf{2s}} 
& \multicolumn{1}{c}{\textbf{3s}} 
& \multicolumn{1}{c}{\textbf{4s}} 
& \multicolumn{1}{c}{\textbf{5s}} &       &  \\
    \midrule
    80:10:10 & \textbf{0.46} & 1.1   & 1.9   & 2.88  & 4.09  & 3.37  & 2.06 \\
    70:10:20 & 0.47  & \textbf{1.08} & \textbf{1.83} & \textbf{2.75} & \textbf{3.89} & \textbf{3.25} & \textbf{2.01} \\
    60:25:15 & 0.5   & 1.2   & 1.98  & 2.92  & 4.07  & 3.34  & 2.13 \\
    50:30:20 & 0.56  & 1.25  & 2.06  & 3.03  & 4.22  & 3.47  & 2.22 \\
    \bottomrule
    \end{tabular}%
  \label{tab:DataSplit}%
\end{table}%

\begin{table*}[b]
  \centering
  \vspace{-10pt}
  \caption{Prediction and Model complexity comparison of DeepTrack with Trajectory forecasting models based on NGSIM dataset.}
    \begin{tabular}{p{4.055em}ccccccccc}
    \toprule
    \multirow{2}[4]{*}{\textbf{Models}} & \multicolumn{5}{c}{\textbf{RMSE horizon (m)}} & \multicolumn{1}{c}{\multirow{2}[4]{*}{\textbf{FDE (m)}}} & \multicolumn{1}{c}{\multirow{2}[4]{*}{\textbf{ADE (m)}}} & \multicolumn{2}{c}{\textbf{Complexity}} \\
\cmidrule{2-6}\cmidrule{9-10}    \multicolumn{1}{c}{} & \multicolumn{1}{c}{\textbf{1 sec}} &
\multicolumn{1}{c}{\textbf{2 sec}} & 
\multicolumn{1}{c}{\textbf{3 sec}} & 
\multicolumn{1}{c}{\textbf{4 sec}} & 
\multicolumn{1}{c}{\textbf{5 secs}} &       &       & \multicolumn{1}{c}{\textbf{MACs}} & \multicolumn{1}{c}{\textbf{Parameters}} \\
    \midrule

    \multicolumn{1}{l}{CS-LSTM \cite{deo2018convolutional}} & 0.61  & 1.27  & 2.09  & 3.10   & 4.37  & 3.34  & 2.29  & 3,580,392 & 191,829 \\
    \multicolumn{1}{l}{CF-LSTM \cite{xie2021congestion}} & 0.55  & 1.1   & 1.78  & 2.73  & 3.82  &  -     & 2.06  & 3,612,494 & 193,941 \\
    \multicolumn{1}{l}{SAAMP \cite{mercat2020multi}} & 0.51  & 1.13  & 1.88  & 2.81  & 3.98  &   -    & 2.06     & \multicolumn{1}{c}{-} & \multicolumn{1}{c}{-} \\
    \multicolumn{1}{l}{STA-LSTM \cite{lin2020attention}} & \textbf{0.37} & \textbf{0.98} & \textbf{1.71} & \textbf{2.63} & \textbf{3.78} & 3.16  & 1.89  & 3,634,456 & 124,835 \\
    DeepTrack~($DT_2$) & 0.47  & 1.08  & 1.83  & 2.75  & 3.89  & 3.25  & 2.01  & {\textbf{2,804,419}} & \textbf{109,099} \\
    \bottomrule
    \end{tabular}%
  \label{tab:model_comparision}%
\end{table*}%


\vspace{-10pt}
\subsection{Comparison Against Existing Approaches}
\label{sec:compared_models}
We compare the results of DeepTrack against the four prominent recently introduced models. (1) Convolutional-social-LSTM (CS-LSTM)~\cite{deo2018convolutional}:  It is based on Social-LSTM~\cite{DBLP:conf/cvpr/AlahiGRRLS16}, an algorithm used for human trajectory detection. CS-LSTM is an encoder-decoder-based model using a social pooling layer to extract the features from the interaction of vehicles in every input sample. (2) CF-LSTM~\cite{xie2021congestion}: A student-teacher network introduced for trajectory prediction. In this network, the LSTM Encoder-Decoder-based model is used for student algorithm and the convolutional graph network for teacher algorithm. (3) Spatiotemporal attention-LSTM (STA-LSTM)~\cite{lin2020attention}: As the name suggests, STA-LSTM uses spatial and temporal information with an attention mechanism to explain the effect of historical trajectories and neighboring vehicles on the ego vehicle. (4) Social Attention Multi-Modal Prediction (SAAMP)~\cite{mercat2020multi}: This model used an LSTM-based encoder-decoder structure with attention layers in the middle to incorporate real-time interactions. It utilizes a multi-head attention mechanism and fuses the long-range attention for joint and multi-modal forecasts.

The performance of DeepTrack and all the models mentioned in section \ref{sec:compared_models} are listed in Table~\ref{tab:model_comparision}. We compare the error and complexity of DeepTrack to state-of-the-art algorithms in vehicle trajectory prediction using NGSIM datasets. As DeepTrack aims to provide best-in-class trajectory forecasting with a low error rate, the following sections analyze the performance in terms of RMSE up to five seconds, FDE, ADE, and complexity of the DeepTrack with other networks.


\subsubsection{Error-Based Analysis}
Compared to CS-LSTM, CF-LSTM, and, SAAMP, DeepTrack can reduces ADE by 12.23\%, 2.43\% and 1.47\% respectively as shown in Table \ref{tab:model_comparision}. DeepTrack also excels for all the steps of RMSE comparison to CS-LSTM and SAAMP. The better performance of DeepTrack is since it has higher gradient stability due to the use of a TCN-based encoder that it is better able to generalize solutions. However, CF-LSTM is better than DeepTrack at 3rd, 4th, and 5th second, and STA-LSTM outperforms DeepTrack. STA-LSTM gives 5.97\% and 2.77\% better ADE and FDE performance than DeepTrack. Thus, DeepTrack under-performs when compared with STA-LSTM with a small margin. 

\subsubsection{Model Complexity Analysis}
DeepTrack outperforms every algorithm in terms of the number of MACs and Parameters, as shown in Table~\ref{tab:model_comparision}. Analyzing and comparing the MAC operations and the size of the model parameters for the approaches mentioned in section~\ref{sec:compared_models}, we anticipated the difference. Deeptrack undercuts STA-LSTM in terms of complexity by 22.84\% fewer MACs count and 12.61\% better parameter count. It also provides 21.67\% better MACs performance, 43.13\% fewer parameters than CS-LSTM, and outperforms CF-LSTM by 22.37\% and 43.75\% in terms of the number of MACs and parameters, respectively. We could not compare its model complexity with DeepTrack and DeepTrack-ATT with SAAMP as the source code was not available publicly.

\begin{figure}[htbp]
    \vspace{-15pt}
    \centering
    \subfigure[Congested traffic]{
    \includegraphics[width=.49\linewidth, trim= 1 1 1 1,clip, keepaspectratio]{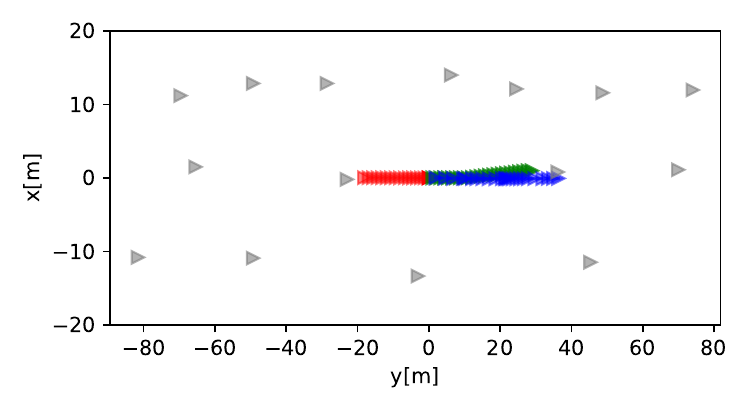}
    \label{fig:congested_tr}
    }
    \subfigure[Lane-keeping]{
    \includegraphics[width=.49\linewidth, trim= 1 1 1 1,clip, keepaspectratio]{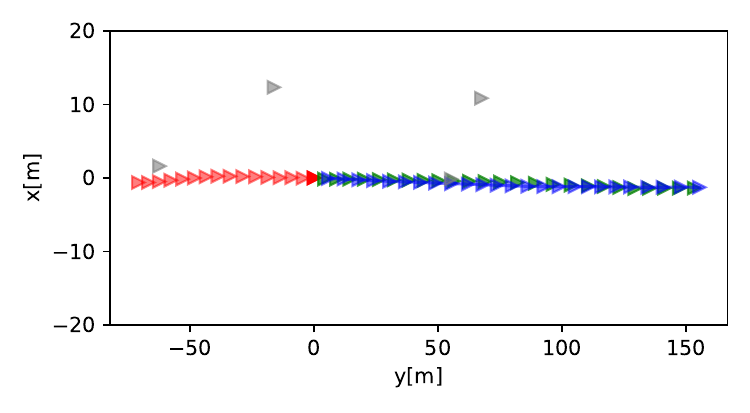}
    \label{fig:lane_keeping}
    }\\
    \subfigure[Maneuvering - Passing from left]{
    \includegraphics[width=.49\linewidth, trim= 1 1 1 1,clip, keepaspectratio]{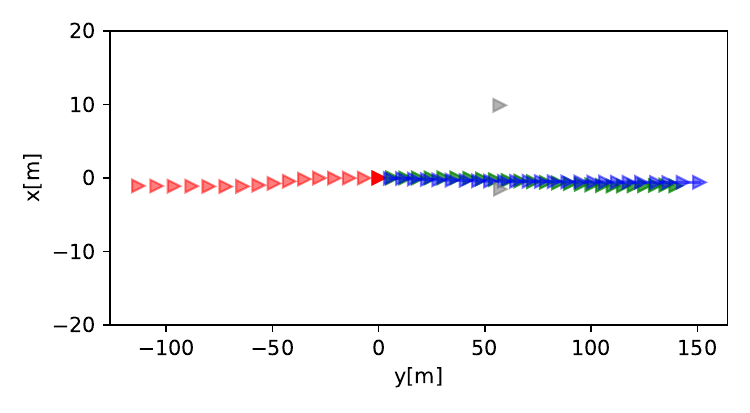}
    \label{fig:manuv_l}
    }
    \subfigure[Maneuvering - Passing from right]{
    \includegraphics[width=.49\linewidth, trim= 1 1 1 1,clip, keepaspectratio]{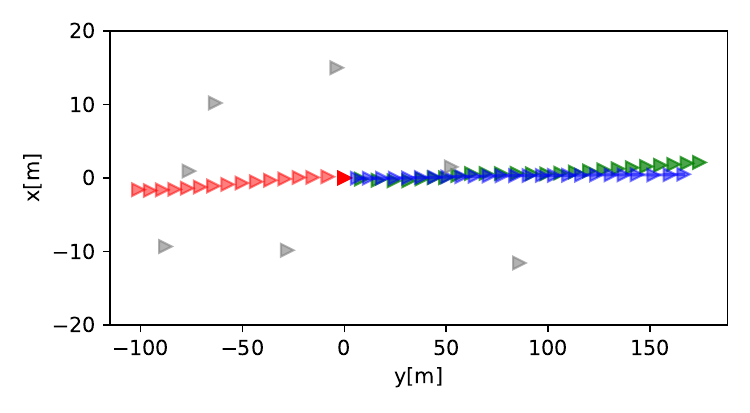}
    \label{fig:manuv_r}
    }

    \caption{The location of the neighbors (gray triangles) is shown at $t_{0}$. Triangles denoting red, green, and blue respectively, represent observed history paths, ground truth, and model output.}
    \vspace{-15pt}
    \label{fig:allGood}
\end{figure}

\begin{figure}[t]
    \centering
    \subfigure[Congested traffic]{
    \includegraphics[width=.49\linewidth, trim= 1 1 1 1,clip, keepaspectratio]{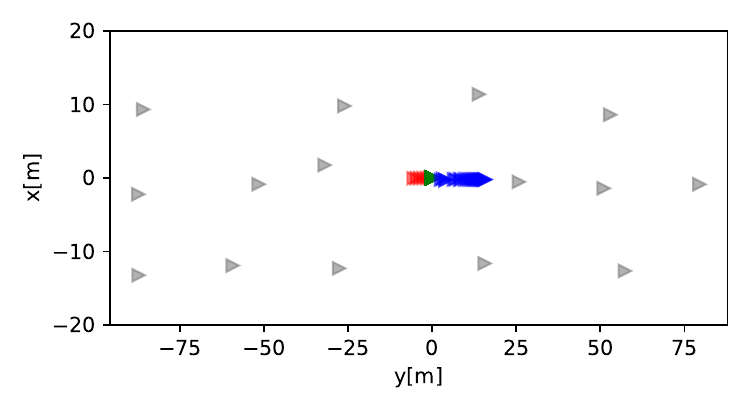}
    \label{fig:cg_bad}
    }
    \subfigure[Multiple lane changing]{
    \includegraphics[width=.49\linewidth, trim= 1 1 1 1,clip, keepaspectratio]{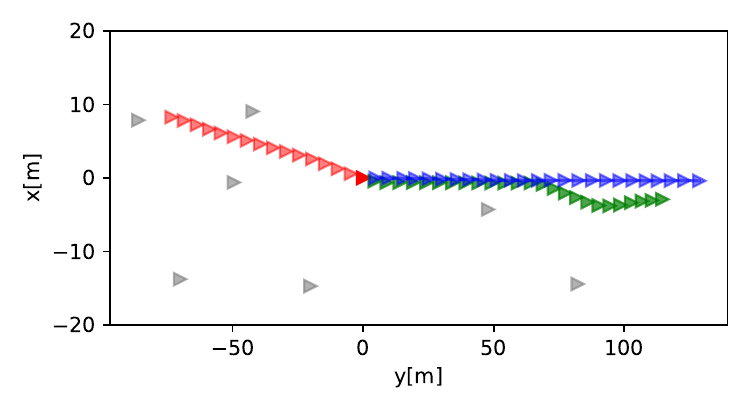}
    \label{fig:ml_bad}
    }
    \caption{Two cases where the model failed to predict the trajectory precisely due to unpredictable driver behaviour. Same legend is used as Fig.~\ref{fig:allGood}.}
        \vspace{-15pt}
    \label{fig:failed}
\end{figure}

\subsection{Qualitative Results}
The analysis of DeepTrack output for different scenarios are discussed in this section.
In Fig. \ref{fig:allGood}-\ref{fig:failed}, the location of the neighbors (gray triangles) are shown at $t_{0}$. Triangles denoting red, green, and blue represent observed history paths, ground truth, and model output. The model predicted output for four scenarios is shown as an aid to understanding the model behaviour: \textcircled{\raisebox{-0.9pt}{1}} congested traffic (Fig.~\ref{fig:congested_tr}), \textcircled{\raisebox{-0.9pt}{2}} lane-keeping (Fig.~\ref{fig:lane_keeping}), \textcircled{\raisebox{-0.9pt}{3}} maneuvering and passing a car from left lane (Fig.~\ref{fig:manuv_l}), \textcircled{\raisebox{-0.9pt}{3}} maneuvering and passing a car from right lane (Fig.~\ref{fig:manuv_r}), and \textcircled{\raisebox{-0.9pt}{4}} cases where the model failed to predict the trajectory precisely (Fig.~\ref{fig:failed}). The three types of the path shown in red, green, and blue triangles represent path history, ground truth, and predicted trajectory for the designated vehicle. The location of the neighbors (gray triangles) is also shown at $t_{0}$. For the sake of simplicity, we did not show the neighbors history path.

The comparison of Fig.~\ref{fig:congested_tr} and Fig. \ref{fig:lane_keeping} shows that the model can accurately estimate the velocity of the interest car based on the ego history. The model correctly predicted that vehicles would travel less distance as a result of congested traffic. The vehicle in the lane-keeping scenario travels farther, and DeepTrack has interfered with the same behavior. Figures \ref{fig:manuv_l} and \ref{fig:manuv_l} illustrate how DeepTrack performs when a car of interest passes its front vehicle from either the left or the right lane. Fig.~\ref{fig:failed} shows the scenarios in which DeepTrack was not able to predict the trajectories due to uncertainty in driver behavior. In the congested scenario (Fig.~\ref{fig:cg_bad}), although the driver slowly drove his car until $t_{0}$, the vehicle stopped for the entire next five seconds, while the model predicts it would come close to the front car.

\vspace{-10pt}
\section{Conclusion}
\label{sec:conclusion}
DeepTrack is a deep learning model with comparable accuracy to best-in-class trajectory prediction algorithms but with a smaller model size and lower computational complexity. The vehicle dynamics are encoded using a TCN-based encoder instead of LSTM units in DeepTrack, and TCN utilizes depthwise convolution, thereby reducing the complexity of models in terms of size and operations compared with LSTMs. The results indicate that DeepTrack reduces the model size and complexity by at least 21.67\%, and 43.13\% compared to CS-LSTM, 22.37\%, and 43.75\% compared to CF-LSTM, and 22.84\%, and 12.61\% than STA-LSTM. The RMSE and displacement errors for DeepTrack are better or comparable to most state-of-the-art trajectory prediction algorithms using NGSIM dataset used in this manuscript.

%

\section*{Acknowledgment}

This work was supported by the National Science Foundation (NSF) under Award No. 1932524.

\bibliographystyle{IEEEtranS}
\bibliography{Bibliography/bibliography.bib}

\vfill

\end{document}